\documentclass{article}

\usepackage{PRIMEarxiv}

\usepackage[utf8]{inputenc} 
\usepackage[T1]{fontenc}    
\usepackage{hyperref}       
\usepackage{url}            
\usepackage{booktabs}       
\usepackage{amsfonts}       
\usepackage{nicefrac}       
\usepackage{microtype}      
\usepackage{lipsum}
\usepackage{fancyhdr}       
\usepackage{graphicx}       
\graphicspath{{media/}}     
\usepackage{amsmath}        
\title{Quantile Extreme Gradient Boosting for Uncertainty Quantification
\thanks{\textit{\underline{Citation}}: 
\textbf{Authors. Title. Pages.... DOI:000000/11111.}} 
}

\author{
  Xiaozhe Yin, Masoud Fallah-Shorshani, Rob McConnell, Scott Fruin \\
  Department of Population and Public Health Sciences, Keck School of Medicine \\
  University of Southern California \\
  Los Angeles, CA 90032 USA\\
  \texttt{\{xiaozhey, shorshan, rmcconne, fruin\}@usc.edu} \\
   \And
  Yao-Yi Chiang \\
  Department of Computer Science \& Engineering \\
  University of Minnesota \\
  Minneapolis, MN 55455 USA\\
  \texttt{yaoyi@umn.edu} \\
   \AND
   Meredith Franklin \\
  Department of Statistical Sciences and School of Environment \\
  University of Toronto \\
  Toronto Ontario, Canada\\
  \texttt{meredith.franklin@utotonto.ca} \\
}

\begin{document}
\maketitle

\begin{abstract}
As the availability, size and complexity of data have increased in recent years, machine learning (ML) techniques have become popular for modeling. Predictions resulting from applying ML models are often used for inference, decision-making, and downstream applications. A crucial yet often overlooked aspect of ML is uncertainty quantification, which can significantly impact how predictions from models are used and interpreted. 

Extreme Gradient Boosting (XGBoost) is one of the most popular ML methods given its simple implementation, fast computation, and sequential learning, which make its predictions highly accurate compared to other methods. However, techniques for uncertainty determination in ML models such as XGBoost have not yet been universally agreed among its varying applications. We propose enhancements to XGBoost whereby a modified quantile regression is used as the objective function to estimate uncertainty (QXGBoost). Specifically, we included the Huber norm in the quantile regression model to construct a differentiable approximation to the quantile regression error function. This key step allows XGBoost, which uses a gradient-based optimization algorithm, to make probabilistic predictions efficiently. 

QXGBoost was applied to create 90\% prediction intervals for one simulated dataset and one real-world environmental dataset of measured traffic noise. Our proposed method had comparable or better performance than the uncertainty estimates generated for regular and quantile light gradient boosting. For both the simulated and traffic noise datasets, the overall performance of the prediction intervals from QXGBoost were better than other models based on coverage width-based criterion. 
\end{abstract}


\section{Introduction}
Modeling complex data such as those arising from environmental processes is challenging given the mix of spatial scales, interactions between components, and underlying spatiotemporal patterns. With the explosion of geospatial data from a variety of sources, including satellites, monitoring networks, and field campaigns, environmental scientists have been adopting data-driven approaches rather than relying solely on more traditional deterministic techniques such as earth system (climate), dispersion, and acoustic modeling. To handle large data sets, artificial intelligence (AI) and machine learning (ML) have become increasingly popular, as simple classification or linear regression methods do not provide the flexibility that allow for non-linearities and complex interactions \cite{Irrgang2021}. Nevertheless, though AI and ML methods such as neural networks and gradient boosting algorithms have achieved high performance in point predictions, they are often still operated as black boxes. More concerningly, predictions from these models have been shown to be miscalibrated, and can yield arbitrarily high confidence \cite{guo2017calibration}; they also usually do not come with any estimation of uncertainty, an important gap. A recent report by the National Academy of Sciences indicated that to make advances at the intersection of AI/ML and Earth system science, understanding and communicating uncertainty is essential for decision-making \cite{NAP26566}. 

Studies have been conducted that quantify uncertainty in deep learning, among which Bayesian methods \cite{kabir2018neural, nourani2021prediction}, the delta method \cite{khosravi2011comprehensive}, bootstrap \cite{zio2006study, errouissi2015bootstrap}, and lower upper bound estimation \cite{nourani2021prediction} are the most commonly used approaches. These uncertainty quantification methods typically give a probability density function as the output, which can be unsatisfactory. For example, some studies argued that the PIs of the bootstrap method were large while coverage was poor (\cite{magnusson2013measuring, zhang2015evaluating, qrfandbootstrap}).  Unfortunately these methods are difficult to generalize to tree-based methods, which is why other approaches need to be developed.

Quantile regression for uncertainty quantification has recently gained popularity in gradient-based optimization algorithms, where it is introduced to construct point-wise prediction intervals (PIs) of specific quantiles. For example, Taylor (2000) \cite{taylor2000quantile} used a quantile regression neural network (QRNN) to estimate the multiple quantiles of multi-period financial returns. QRNN has also been used in predicting wind power \cite{he2018probability}, electricity consumption \cite{he2019electricity}, rainfall extremes \cite{cannon2018non}, and to evaluate online teaching \cite{pan2016study}. Xu et al. (2017) \cite{xu2017composite} developed a composite quantile regression neural network and applied it to three standard datasets, showing that it had better performance than QRNN.  \par

Quantile regression has a solid statistical foundation and can be easily customized and incorporated into the cost function of many ML models. For example, quantile regression random forest (QRRF) has been developed, and some studies have suggested that this method is competitive in terms of its predictive power \cite{meinshausen2006quantile}. There have been applications of QRRF in statistical load forecasting \cite{aprillia2020statistical}, uncertainty estimation of digital soil mapping products \cite{vaysse2017using} and suspended sediment concentration \cite{francke2008estimation}. Quantile regression has also been implemented in gradient-boosting-tree-based methods, such as gradient boosting machine (GBM) \cite{natekin2013gradient} and light gradient boosting machine (LightGBM). Quantile GBM has been used in environmental probabilistic forecasting problems, including to forecast wind power \cite{landry2016probabilistic} and solar irradiation \cite{verbois2018probabilistic}. Both studies found that including quantiles significantly improved the models by enabling prediction of quantiles for uncertainty quantification. As an emerging machine learning method, LightGBM has been applied to predict particulate matter air pollution from satellite observations \cite{wei2021himawari}. However, we have not seen an application of the quantile LightGBM. \par

Despite quantile regression gaining popularity in neural networks and some tree-based machine learning methods, it has never been used in extreme gradient boosting (XGBoost) for two reasons. First, the quantile regression function is not differentiable at 0, meaning that the gradient-based XGBoost method might not converge properly and lead to sub-optimal model parameters \cite{guitton2003robust}. Second, different from GBM, a second-order approximation is required to optimize the regularized objective function in XGBoost, and the second derivative of the quantile regression function is 0 everywhere. To solve these limitations, we incorporated the use of the Huber norm in combining the quantile regression method with XGBoost and we tested the proposed quantile XGBoost method in both a simulated and a real-world traffic noise dataset. Even though we used environmental data as an example in this study given the complexity in geospatial data, QXGBoost can be universally applied to any type of data for which uncertainty analysis is desired.  \par

\section{Extreme Gradient boosting}\label{sec2}
Built upon the widely used tree-based boosting method, XGBoost was proposed by Chen and Guestrin (2016) \cite{chen2016xgboost} to use additive linear functions to predict the output as shown in equation \ref{eq:base learner}.
\begin{equation}
    \label{eq:base learner}
    \hat{y_i}=\phi(x_i)=\sum_{k=1}^k f_k(x_i), ~ f_k \in \mathcal{F}  
\end{equation}
where $\mathcal{F}=\{f(x)=w_{q(x)}\} (q: \mathbb{R}^m \to T, w \in \mathbb{R}^T)$ is the space of regression trees and $q$ is the distinctive structure of each tree that is represented by $f(k)$ to map an input to the corresponding leaf index; $K$ is the number of trees and $T$ is the number of leaves in each tree; $w$ represents the leaf weights (scores). The final prediction of a specific input is calculated by summing up the scores ($w$) in the corresponding leaves. \par 
As with all supervised machine learning models, XGBoost uses an objective function to train the data and assess the model fit. Instead of using just the loss function, XGBoost adds an additional regularization term to prevent overfitting. For regression problems, mean squared error is a common choice for loss function and the corresponding objective function can be expressed as:
\begin{equation}
\begin{aligned}
    \label{eq:objective function}
    \mathcal{L}(\theta)^{(t)}&=L(\theta)+\Omega(\theta)\\
                           &=\sum_i l(\hat{y_i}, y_i)+\sum_k \Omega(f_k)\\
                           &=\sum_{i=1}^n (y_i - (\hat{y_i}^{(t-1)}+f_t (x_i)))^2+\Omega(f_t)
\end{aligned}                           
\end{equation}
Here $L$ represents the differentiable convex loss function that measures the difference between the prediction $\hat{y_i}$ and the target $y_i$. The second term $\Omega$ is the regularization term that penalizes the complexity of the model, and $\Omega(f)=\gamma T+\frac{1}{2}\lambda w^2$, where $T$ refers to the number of leaves in each tree; $\gamma$ and $\lambda$ are the $L1$ and $L2$ regularization terms that control weights at leaves.\par
The loss function for regression problems can easily be calculated using first-order and quadratic terms (equation \ref{eq:objective function}). However, the loss functions for other losses of interest, such as the logistic loss, are not friendly. To overcome this issue, XGBoost employs the Taylor expansion, originally proposed by Friedman et al. (2000) \cite{friedman2000additive} to quickly optimize the objective function in equation \ref{eq:objective function} to the second order. The Taylor expansion of function $f(x)$ that is infinitely differentiable at a real number $a$ can be written as: 
\begin{equation}
\begin{aligned}
    \label{eq:taylor expansion}
     f(x)&=f(a) + \frac{f^{'}(a)}{1!}(x-a)+ \frac{f^{''}(a)}{2!}(x-a)^2+\frac{f^{'''}(a)}{3!}(x-a)^3+...\\
         &=\sum_{n=0}^{\infty} \frac{f^{(n)}(a)}{n!}(x-a)^n
\end{aligned}
\end{equation}
In practice, the higher orders of the Taylor expansion function are usually omitted to get the approximation of the function. XGBoost keeps the constant, first, and second derivatives of $f(x)$ evaluated at $a$. By replacing $f(x)$ with $(f_t(x_i)+\hat{y_i}^{(t-1)})-y_i$ and expanding $(f_t(x_i)+\hat{y_i}^{(t-1)})-y_i$ at $(\hat{y_i}^{(t-1)})-y_i)$ in equation \ref{eq:objective function}, the objective function can be rewritten as:
\begin{equation}
    \label{eq:taylor expansion2}
     \mathcal{L}^{(t)} \simeq \sum_{i=1}^n [l(y_i, \hat{y_i}^{(t-1)}) + g_i{f_t (x_i)}+ \frac{1}{2}h_if_t^2(x_i)]+\Omega(f_t)
\end{equation}
where the $g_i$ and $h_i$ are the first and second order gradient statistics of the loss function and they are defined as 
\begin{equation}
\begin{aligned}
    \label{eq:g_i and h_i}
     &g_i=\partial_{\hat{y_i}^{(t-1)}}l(y_i, \hat{y_i}^{(t-1)})\\
     &h_i=\partial_{\hat{y_i}^{(t-1)}}^2l(y_i, \hat{y_i}^{(t-1)})\\
\end{aligned}
\end{equation}
In XGBoost, only the functions $f_t(x_i)$ are treated as parameters. Thus, the $l(y_i, \hat{y_i}^{(t-1)})$ can be considered as constant terms. Given that all the instances on the same leaf get the same score, if we define $I_j=\{i|q(x_i)=j\}$ as the set of indices of data points assigned to the $j$-th leaf, $G_i=\sum_{i\in I_j}g_i$ and $H_i=\sum_{i\in I_j}h_i$, and substitute the regularization term $\Omega(f_t)$, the objective function of equation \ref{eq:g_i and h_i} can be further compressed as 
\begin{equation}
\begin{aligned}
    \label{eq:objective function5}
     \mathcal{\tilde{L}}^{(t)} &\simeq \sum_{i=1}^n [g_i{f_t (x_i)}+ \frac{1}{2}h_if_t^2(x_i)]+\gamma T+\frac{1}{2}\lambda w^2\\
     &=\sum_{j=1}^T[(\sum_{i \in I_j}g_i)w_j+\frac{1}{2}(\sum_{i \in I_j}h_i + \lambda)w_j^2]+\gamma T\\
     &=\sum_{j=1}^T [G_j w_j+ \frac{1}{2}(H_j+\lambda)w_j^2]+\lambda T
\end{aligned}
\end{equation}
For a fixed structure, the optimal weight $w_j^*$ of leaf $j$ can be obtained by 
\begin{equation}
    \label{eq:optimal weights}
     w_j^*=-\frac{G_j}{H_j+\lambda}
\end{equation}
The corresponding objective function when plugging $w_j^*$ back to equation \ref{eq:objective function5} is
\begin{equation}
    \label{eq:optimal objective function value}
     \mathcal{\tilde{L}}^{(t)} = -\frac{1}{2} \sum_{j=1}^T \frac{G_j^2}{H_j+\lambda}+\lambda T
\end{equation}
In practice, function \ref{eq:optimal objective function value} was used to evaluate the tree structure. The smaller the score, the better the tree structure. From equation \ref{eq:optimal objective function value}, we can see that the value of the objective function only depends on $g_i$ and $h_i$, meaning that any function with $g_i$ and $h_i$ can be used as a potential objective function. Thus, XGBoost supports the customized objective function. XGBoost uses a greedy algorithm to search for the best split of a tree. Specifically, the following score gain function is used to split a leaf into two leaves
\begin{equation}
    \label{eq:optimal split function}
     \mathcal{L}_{splitgain} 
     =\frac{1}{2}[\frac{G_L^2}{H_L+\lambda}+\frac{G_R^2}{H_R+\lambda}-\frac{(G_L+G_R)^2}{H_L+H_R+\lambda}]-\gamma
\end{equation}
where $I_L$ and $I_R$ are the instance sets of the new left and right leaf after the split and $I=I_L+I_R$
is the score on the original leaf. The larger the split score, the better the split. The best split of a leaf happens when the split gain score is the largest among all possible split candidates. For equation \ref{eq:optimal split function}, if the split gain scores are all smaller than $\gamma$, which means that the split score is negative, then there will be no split at that leaf. Equation \ref{eq:optimal split function} is usually used in practice to evaluate all split candidates. 

\section{Quantile regression}\label{sec3}
\begin{figure}[!ht]
\centerline{\includegraphics[width=.6\linewidth,height=.5\textheight,keepaspectratio]{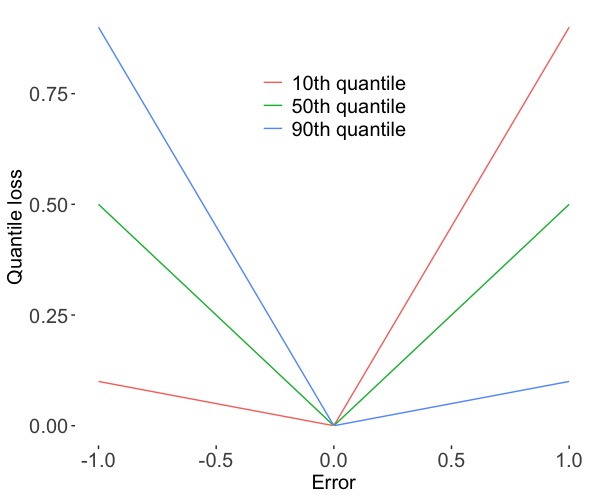}}
\caption{Quantile plot.\label{fig:quantile plot}}
\end{figure}

Quantile regression was first introduced by Koenker and Bassett Jr (1978) \cite{koenker1978regression} to estimate and conduct inference about conditional quantile functions. Instead of minimizing the least square, quantile regression minimizes the mean absolute error, which applies asymmetric weights to positive and negative errors to get the entire conditional distribution \cite{koenker1978regression}. Thus, quantile regression is the process of changing the mean square error loss function to one that predicts conditional quantiles rather than conditional means \cite{yu2003quantile}. The absolute value function, which is also called the check, tick, or pinball loss function, is given by
\begin{equation}
\label{eq:quantile function}
\rho_\tau(x) = 
\begin{cases}
     (\tau-1)x & x<0\\
     \tau x & otherwise
\end{cases}
\end{equation}
where $\tau$ is a specific quantile that we want to calculate ($0<\tau<1$) and $x$ represents the prediction error $y_i-\hat{y_i}$. Given that the mean absolute error is $\frac{1}{n}\sum_{i=1}^n|y_i-\hat{y_i}|$, the corresponding mean quantile loss can be expressed as $\frac{1}{n}\sum_{i=1}^n \rho_q (y_i-\hat{y_i})$. To prove that the optimal estimate of this loss function is actually the quantiles, we calculate the expectation of the absolute value function as
\begin{equation}
\label{eq:quantile function prove}
\begin{aligned}
     E[\rho_\tau (y-\hat{y})] &=(\tau-1)\int_{-\infty}^{\hat{y}} f(t)(t-\hat{y})dt+\tau \int_{\hat{y}}^{\infty} f(t)(t-\hat{y})dt\\
     &=\tau \int f(t)(t-\hat{y})dt -\int_{-\infty}^{\hat{y}} f(t)(t-\hat{y})dt\\
     &=\tau \int f(t)tdt - \hat{y}\tau \int f(t)dt - \int_{-\infty}^{\hat{y}}f(t)tdt + \hat{y}\int_{-\infty}^{\hat{y_i}}f(t)dt\\
     &=\tau \int f(t)tdt - \hat{y}\tau - \int_{-\infty}^{\hat{y}}f(t)tdt + \hat{y}F(\hat{y})
\end{aligned}
\end{equation}
Here $f$ and $F$ represent the probability density and the cumulative distribution functions of $Y$, respectively. If we differentiate the equation \ref{eq:quantile function prove} in terms of $\hat{y}$ and set the expression equal to 0, we get 
\begin{equation}
\label{eq:quantile function prove result}
\begin{aligned}
     \frac{\partial E[\rho_\tau (y-\hat{y})]}{\partial \hat{y}}=0
     &=\frac{\partial (\tau \int f(t)tdt - \hat{y}\tau - \int_{-\infty}^{\hat{y}}f(t)tdt + \hat{y}F(\hat{y}))}{\partial \hat{y}}\\
     &=\frac{\partial (\hat{y}F(\hat{y}))}{\partial \hat{y}} + \frac{\partial (\tau \int f(t)tdt)}{\partial \hat{y}} - \frac{\partial (\hat{y}\tau)}{\partial \hat{y}} - \frac{\partial \int_{-\infty}^{\hat{y}}f(t)tdt}{\partial \hat{y}}\\
     &=\frac{F(\hat{y}) \partial \hat{y}}{\partial \hat{y}} + \frac{\hat{y} \partial F(\hat{y})}{\partial \hat{y}} + 0 - \tau \frac{\partial \hat{y}}{\partial \hat{y}} - \hat{y}f(\hat{y})\\
     &=F(\hat{y})+\hat{y}f(\hat{y})-\tau-\hat{y}f(\hat{y})\\
     &=F(\hat{y})-\tau
\end{aligned}
\end{equation}
Solving the above equation with respect to $\hat{y}$, we get $\hat{y}=F^{-1} (\tau)$, which is the specific quantile that we want to calculate. Equation \ref{eq:quantile function prove result} proves that the prediction $\hat{y}$ is the $\tau^{th}$ quantile. The absolute value function is a non-negative function as shown in Figure \ref{fig:quantile plot}


\section{Quantile extreme gradient boosting}
The proposed quantile extreme gradient boosting (QXGBoost) method combines quantile regression and XGBoost to construct prediction intervals (PIs). Specifically, instead of using the mean square error as the loss function in XGBoost for regression problems, QXGBoost employs the absolute value function as the loss function to calculate the PIs, which quantify uncertainties. By doing so, QXGBoost can create specific prediction quantiles that can be customized depending on the application area. \par
In XGBoost, the optimal weights (equation \ref{eq:optimal weights}), structure score (equation \ref{eq:optimal objective function value}), and the split gain function (equation \ref{eq:optimal split function}) all depend on the gradient statistics of each instance. Thus, the objective function in XGBoost must meet two requirements. First, it must be continuously differentiable in the defined sample space. Second, the first and second derivatives of the loss function should exist and not be 0 everywhere. However, the absolute value function (equation \ref{eq:quantile function}) is not differentiable at the origin since it has a ``break'' point and the second derivative is 0 everywhere (figure \ref{fig:quantile plot}). Thus, the original quantile regression function cannot be used directly as the customized objective function in XGBoost. \par
\begin{figure}
\centering
  \includegraphics[width=.6\linewidth,height=.5\textheight,keepaspectratio]{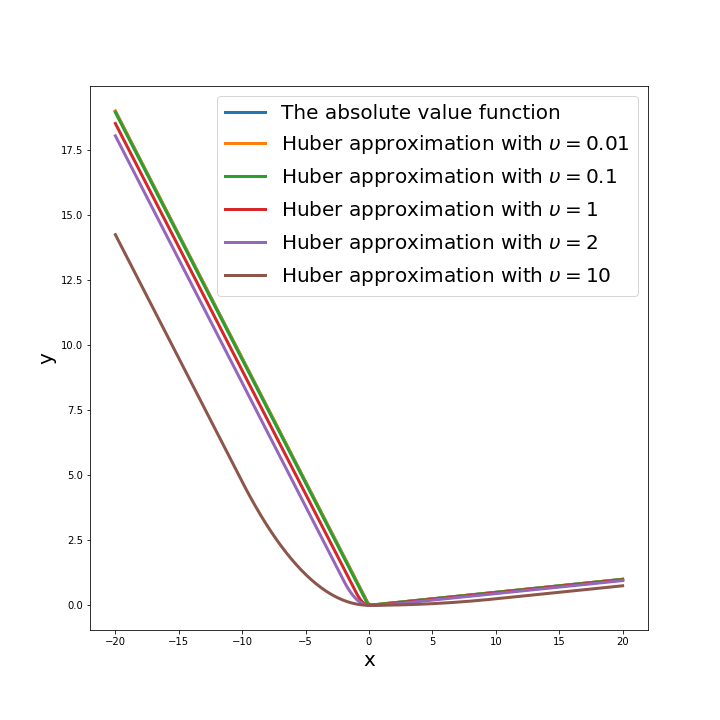}
  \caption{The difference between the absolute value function and Huber approximation function with different thresholds when the quantile $\tau=0.95$.}
  \label{fig:huber plot}
\end{figure}

To solve the two problems imposed for implementing the quantile regression in XGBoost, we need a differentiable approximation to the quantile regression error function so that the gradient-based optimization algorithm can work. We followed the work by Chen (2007) \cite{chen2007finite}, who introduced the Huber function \cite{huber1973robust} to construct the smooth approximation of the absolute value function and thus has a smooth transition at 0. The Huber norm function has been used as a loss function in neural networks \cite{cannon2011quantile}, but this is the first use of it in another algorithm, including XGBoost. The Huber function is written as:
\begin{equation}
\label{eq:huber function}
h_\upsilon (t) = 
\begin{cases}
     \frac{t^2}{2\upsilon} & |t| \le \upsilon\\
     |t|-\frac{\upsilon}{2} & |t|>\upsilon
\end{cases}
\end{equation}
where the threshold $\upsilon$ is a positive real number. By replacing the quantile error function $x$ in the absolute value function \ref{eq:quantile function} with the Huber function \ref{eq:huber function}, we get the new objective function and its first and second derivatives as follows:
\begin{equation}
\label{eq:huber quantile function}
\begin{aligned}[t]
 \rho_\tau = 
\begin{cases}
     (\tau-1)(|t|-\frac{\upsilon}{2}) & t < -\upsilon\\
     (\tau-1)(\frac{t^2}{2\upsilon}) & -\upsilon \le t <0\\
     \tau (\frac{t^2}{2\upsilon}) & 0 \le t < \upsilon\\
     \tau(|t|-\frac{\upsilon}{2}) & t>\upsilon
\end{cases}
\end{aligned}
\quad\text{,}\quad
\begin{aligned}[t]
 \nabla_\rho = 
\begin{cases}
     (1- \tau) & t < -\upsilon\\
     (\tau-1)(\frac{t}{\upsilon}) & -\upsilon \le t <0\\
     \tau (\frac{t}{\upsilon}) & 0 \le t < \upsilon\\
     \tau & t>\upsilon
\end{cases}
\end{aligned}
\quad\text{, and}\quad
\begin{aligned}[t]
 \nabla_{\rho}^2 = 
\begin{cases}
     0 & t < -\upsilon\\
     \frac{\tau-1}{\upsilon} & -\upsilon \le t <0\\
     \frac{\tau}{\upsilon} & 0 \le t < \upsilon\\
     0 & t>\upsilon
\end{cases}
\end{aligned}
\end{equation}
Our proposed QXGBoost method uses equation \ref{eq:huber quantile function} as the objective function for XGBoost to construct PIs. Equation \ref{eq:huber quantile function} proves that the proposed objective function is differentiable in the defined space, and the second derivative exists and is not 0 everywhere. Figure \ref{fig:huber plot} compares the original absolute value function and the Huber smoothed absolute value functions with different thresholds when the quantile $\tau=0.95$. As we can see, the Huber smoothed absolute value function $h_\upsilon (t)$ is continuously differentiable and provides a smooth transition between absolute and square errors around the origin \cite{cannon2011quantile}. When the threshold ($\upsilon$) is smaller than 1, the Huber smoothed absolute value functions approach the absolute value function infinitely. As the threshold $\tau$ gets larger, the Huber smoothed absolute value functions depart further away from the original absolute value function. Thus, in practice, we prefer to use smaller threshold $\tau$ values. \par

In summary, inspired by the idea of quantile neural networks, GBM, LightGBM, and random forest, we came up with the idea of combining quantile regression with XGBoost for the latter to do uncertainty quantification. For the quantile regression function to be a qualified objective function (have first and second derivatives everywhere in the defined sample space and not all 0), we introduced the Huber smoothed absolute value function, which is a differentiable approximation to the absolute value function that provides a smooth transition between absolute and squared errors around the origin. Accordingly, two sets of parameters need to be tuned for QXGBoost. One set of parameters is inhabited in the XGBoost model, the most important parameters of which include the number of estimators, the maximum tree depth, and the learning rate. The other set comes with the threshold parameter in the Huber function ($\upsilon$) inside the smoothed quantile regression function. Both sets of parameters need to be tuned to achieve the desired model performance. 

\section{Evaluation of prediction intervals}
The evaluation of traditional point predictions usually adopts statistical methods, such as root mean square error, mean absolute error, and mean absolute percentage error. However, the evaluation of PIs is not as straightforward and no established approach exists. In practice, the most commonly used criteria for quantifying PIs are the prediction interval coverage probability (PICP) and prediction interval average width (PIAW) \cite{kabir2018neural, nourani2021prediction, pearce2018high, lai2021exploring}. \par
PICP measures the percentage of the target data points that fall in the bounds of the lower and upper limits of the PIs inclusive. The mathematical expression of PICP is represented in equation \ref{eq:PICP}
\begin{equation}
    \label{eq:PICP}
     PICP=\frac{1}{n}\sum_{i=1}^n c_i
\end{equation}
where
\begin{equation}
c_i=
\begin{cases}
    \label{eq:PICP_c}
     1 & t_i \in [\underbar{y}_i, \bar{y}_i]\\
     0 & otherwise
\end{cases}
\end{equation}
Here, $n$ represents the number of samples. $t_i$ is the true $i^{th}$ target value. $\underbar{y}_i$ and $\bar{y}_i$ are the lower and upper bound of the $i^{th}$ sample, individually. Based on equation \ref{eq:PICP}, PICP depends on the width of PIs. Usually wider width of PIs means a larger PICP. However, a very large (for example, 100\%) is meaningless as it conveys no information about the target values. \par 
PIAW calculates the average width of PIs. Given that different datasets usually have different sample ranges, PIAW needs to be normalized so that it can be used to compare across different datasets. The normalized PIAW (PINAW) is given as 
\begin{equation}
    \label{eq:PIAW}
     PINAW=\frac{1}{R \times n}\sum_{j=1}^n (\bar{y}_i - \underbar{y}_i)
\end{equation}
where $R$ is the range of the observed data. PICP and PINAW are positively correlated and typically, a large PICP with small PINAW is preferred. Thus, PICP and PINAW should be used together to co-evaluate the quality of PIs. In practice, we could control one criterion and compare the other criteria among different models. For example, to compare PIs from different models, we could try to set the PIAPs all the same and compare PINAWs. The smaller PINAW is, the better the model is. However, most of the time, Both PICP and PINAW vary among models. To solve the dilemma of PICP and PINAW going in opposite directions, a new criterion called the coverage width-based criterion (CWC), which combines both PICP and PINAW is proposed \cite{khosravi2011comprehensive}
\begin{equation}
CWC=
\begin{cases}
    \label{eq:cwc}
     PINAW(1+e^{\eta(\tau-PICP)}) & PICP<\tau\\
     PINAW & PICP \ge \tau
\end{cases}
\end{equation}
where $\eta=50$ is a hyperparameter and $\tau$ is the desired quantile \cite{kabir2018neural}. CWC is used in practice to assess the quality of PIs across different models. All three methods were adopted to quantify the PIs in the following applications.

\section{Application to simulated and real-world data}
In this section, we applied QXGBoost on simulated data and real-world traffic noise data. To evaluate the performance of QXGBoost, we compared the quality of PIs from QXGBoost with two other uncertainty quantification methods, quantile GBM, and quantile LightGBM. For each method, 90\% PIs were obtained using two models, one predicting the upper bound of PIs with $\tau=0.95$ and one for the lower bound with $\tau=0.05$. Data were randomly split and all models were trained on 75\% of the data and tested on the rest 25\%. \par

\subsection{Simulated data}
The simulation data that we chose was adapted from the Scikit-learn library example which shows how quantile regression can be used to create PIs for GBM and it was generated using the equation \ref{eq:toy data}
\begin{equation}
    \label{eq:toy data}
     y=1.5\times x \times sin(x) + \varepsilon
\end{equation}
where $\varepsilon$ has a normal distribution with mean 0 and standard deviation that was randomly drawn between 1.5 and 2.5 so that we have heteroscedastic noise for the simulated data. The relationship between $x$ and $y$ is not linear. \par
The hyperparameters tuning is shown in table \ref{tab:Hyperparameters tuning for the simulated data}. Quantile GBM and LightGBM shared the same tree-based parameters tuning. QXGBoost used a larger number of trees and maximum tree depth. All three methods used the same learning rate (0.05) for the simulated data. The extra threshold parameter $\tau$ in QXGBoost was set to 2, meaning that the smoothed loss function approximates well to the original absolute value function (figure \ref{fig:huber plot}). \par 
\begin{center}
\begin{table}
\centering
\caption{Hyperparameters tuning for the simulated data.\label{tab:Hyperparameters tuning for the simulated data}}%
\begin{tabular*}{500pt}{@{\extracolsep\fill}lcccc@{\extracolsep\fill}}
\toprule
\textbf{Method} & \textbf{learning rate}  & \textbf{Number of estimators}  & \textbf{Max depth}  & \textbf{Huber threshold} \\
\midrule
Quantile GBM & 0.05 & 200 & 2  & --   \\
Quantile LightGBM & 0.05  & 200  & 2  & --  \\
QXGBoost & 0.05  & 300  & 3  &  2  \\
\bottomrule
\end{tabular*}
\end{table}
\end{center}

\begin{figure}
\centering
  \includegraphics[width=.8\linewidth,height=.7\textheight,keepaspectratio]{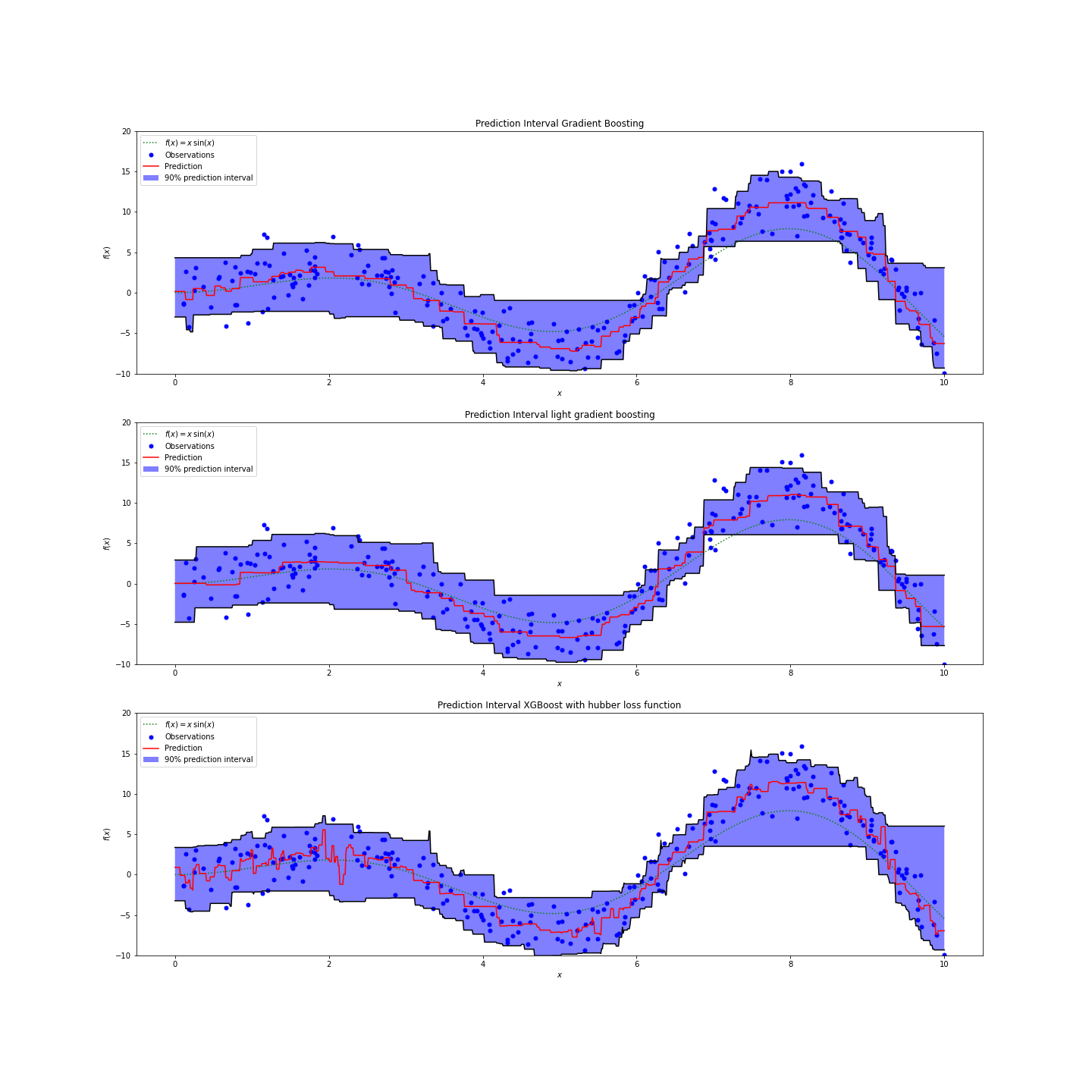}
  \caption{Comparison of prediction intervals for the simulated data using three methods. The blue dots is the observations. The red line is the point prediction of the traffic noise. The two black lines represent the upper and lower bounds of PIs and the blue areas in between are the PIs.}
  \label{fig:toy_pi}
\end{figure}

Table \ref{tab:Prediction interval results for the simulated data} shows that even though QXGBoost has the smallest training PICP (0.891), it has the largest testing PICP (0.892), meaning that QXGBoost does not overfit and can be generalized to other datasets better than quantile LightGBM and quantile GBM. The higher PICP of QXGBoost also comes with a slightly larger PINAW (0.777), which is unsurprising. Overall, QXGBoost outperformed quantile GBM and LightGBM with the smallest CWC (1.937 VS 3.704 and 3.682). Figure \ref{fig:toy_pi} shows that all three models have roughly the same PIs when $x<7$. QXGBoost has smaller values for the lower quantile and higher values for the upper quantile when $7<x<9$ and $9<x<10$, respectively. \par
However, this is a very simple example with only one independent variable, $x$. For XGBoost, this means that there is only one variable used to split and build all the trees. The strong suit of ML methods is their capability to untangle the complex non-linear relationship among the independent variables. Thus, a more complicated example is illustrated using a real-life environmental dataset. \par
\begin{center}
\begin{table}
\centering
\caption{Prediction intervals results for the traffic noise data.\label{tab:Prediction interval results for the simulated data}}%
\begin{tabular*}{500pt}{@{\extracolsep\fill}lcccc@{\extracolsep\fill}}
\toprule
\textbf{Method} & \textbf{Training PICP}  & \textbf{Testing PICP}  & \textbf{testing PINAW}  & \textbf{Testing CWC} \\
\midrule
Quantile GBM & 0.904 & 0.872 & 0.733  & 3.704   \\
Quantile LightGBM & 0.901  & 0.872  & 0.728  & 3.682  \\
QXGBoost & 0.891  & 0.892  & 0.777  &  1.937  \\
\bottomrule
\end{tabular*}
\end{table}
\end{center}

\subsection{Traffic noise data}
Traffic noise data came from Yin et al. (2020) \cite{yin2020predicting}. In brief, traffic noise was measured while walking along 16 predefined routes in Long Beach, CA with two hand-held Brüel \& Kjær 2250 sound analyzers, which are high-end sound meters that allow the user to measure A-weighted, equivalent noise (LAeq) at intervals as short as one second. The sound meters were coupled with Strava to get real-time GPS location data. After data processing, a total of 6,647 geo-referenced observations were obtained. Based on their latitude and longitude, the 44 predictor variables, including meteorology, traffic, land use, and road features, were extracted to build and train the models. \par
\begin{figure}[!ht]
\includegraphics[width=1\textwidth]{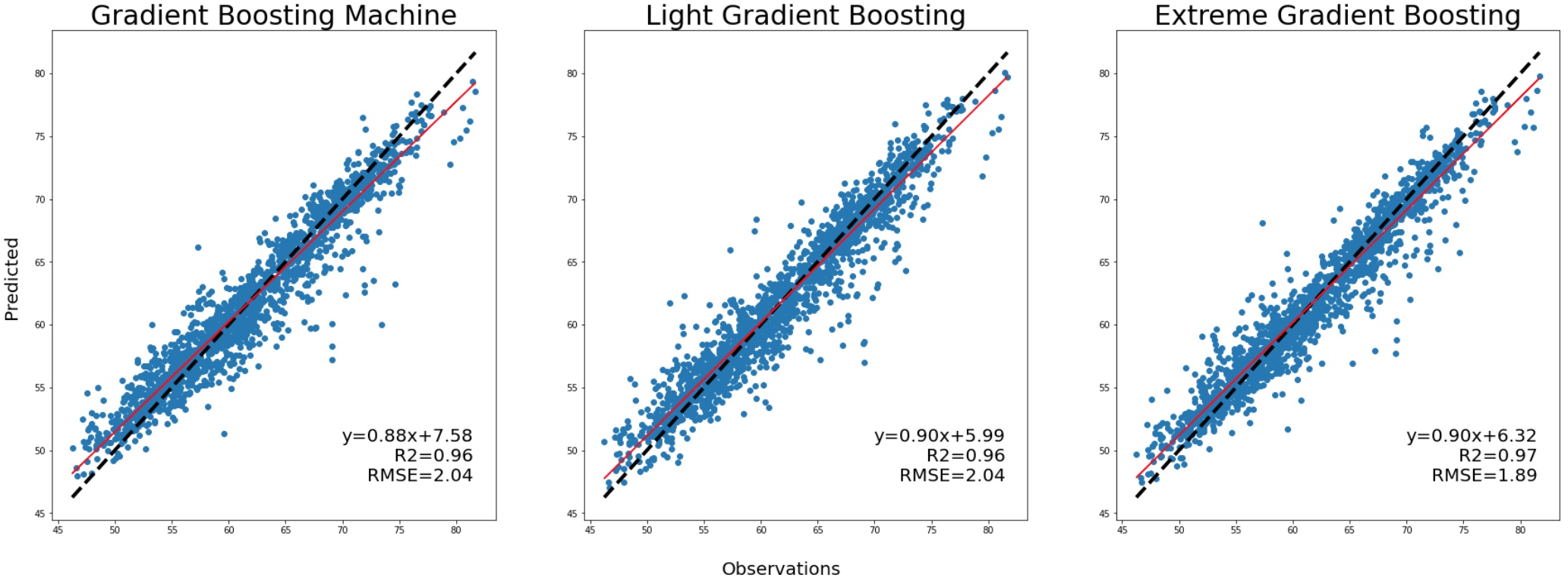}
\centering\caption{Point prediction for traffic noise using GBM, LightGBM, and XGBoost. Blue points represent each observed and predicted sample; solid red lines represent the regression line; and dashed black lines represent the 1:1 correspondence between measured and predicted values.}\label{fig:point prediction for noise using 3 models}
\end{figure}

Before performing the uncertainty quantification analysis, we first run the point prediction analysis of traffic noise using GBM, LightGBM, and XGBoost. The parameters of each algorithm were tuned by a 5-fold CV random grid search process that included measured average LAeq from all of the observations together. After tuning, each model was trained on 70\% of the data and predicted on the remaining 30\%; the prediction results are shown in Figure \ref{fig:point prediction for noise using 3 models}. GBM and LightGBM have very similar prediction performance with the same R\textsuperscript{2} = 0.96 and RMSE = 2.04. XGBoost has a better model performance with a slightly higher R\textsuperscript{2} = 0.97 and a lower RMSE = 1.89.

The tuned hyperparameters for Quantile GBM, Quantile Light GBM, and QXGBoost are shown in table \ref{tab:Hyperparameters tuning for the traffic noise data}. Quantile GBM and LightGBM have the same tree-based tuning parameters. By comparison, QXGBoost used more trees (300 vs 200) to build the models but also a higher learning rate (0.05 vs 0.03). The Huber threshold for the traffic noise data was tuned at 0.07, which gave us a very close approximation to the real quantile regression function according to figure \ref{fig:huber plot}.

\begin{center}
\begin{table}[t]%
\centering
\caption{Hyperparameters tuning for the traffic noise data.\label{tab:Hyperparameters tuning for the traffic noise data}}%
\begin{tabular*}{500pt}{@{\extracolsep\fill}lcccc@{\extracolsep\fill}}
\toprule
\textbf{Method} & \textbf{learning rate}  & \textbf{Number of estimators}  & \textbf{Max depth}  & \textbf{Huber threshold} \\
\midrule
Quantile GBM & 0.03 & 200 & 3  & --   \\
Quantile LightGBM & 0.03  & 200  & 3  & --  \\
QXGBoost & 0.05  & 300  & 3  &  0.07  \\
\bottomrule
\end{tabular*}
\end{table}
\end{center}

Table \ref{tab:Prediction interval results for traffic noise data} summarizes the training PICP and the testing PICP, PINAW, and CWC from all three models. Among them, quantile LightGBM has the largest training PICP (0.907) and the smallest testing PICP (0.885), indicating that the LightGBM might have an overfitting problem. Even though quantile GBM has the largest testing PICP (0.890), it also has the biggest PI width (0.331). By comparison, QXGBoost has a reasonable testing PICP (0.889) with the smallest PINAW (0.319). Overall, QXGBoost has the smallest testing CWC (4.580), followed by quantile LightGBM and quantile GBM (4.610 and 4.751, respectively). \par
\begin{center}
\begin{table}[t]%
\centering
\caption{Prediction intervals results for the traffic noise data.\label{tab:Prediction interval results for traffic noise data}}%
\begin{tabular*}{500pt}{@{\extracolsep\fill}lcccc@{\extracolsep\fill}}
\toprule
\textbf{Method} & \textbf{Training PICP}  & \textbf{Testing PICP}  & \textbf{testing PINAW}  & \textbf{Testing CWC} \\
\midrule
Quantile GBM & 0.905 & 0.890 & 0.331  & 4.751   \\
Quantile LightGBM & 0.907  & 0.885  & 0.320  & 4.610  \\
QXGBoost & 0.902  & 0.889  & 0.319  &  4.580  \\
\bottomrule
\end{tabular*}
\end{table}
\end{center}

\begin{figure}
    \centering
    \includegraphics[width=1\textwidth]{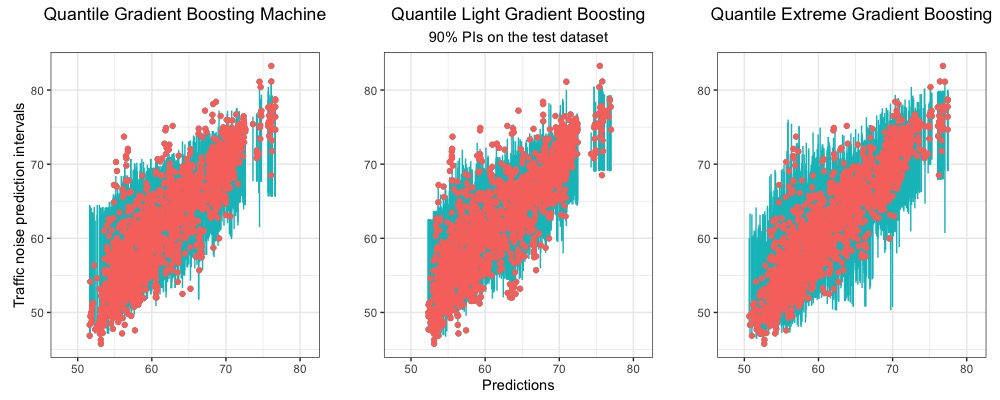}
    \label{fig:pi_noise_r}
    \includegraphics[width=1\textwidth]{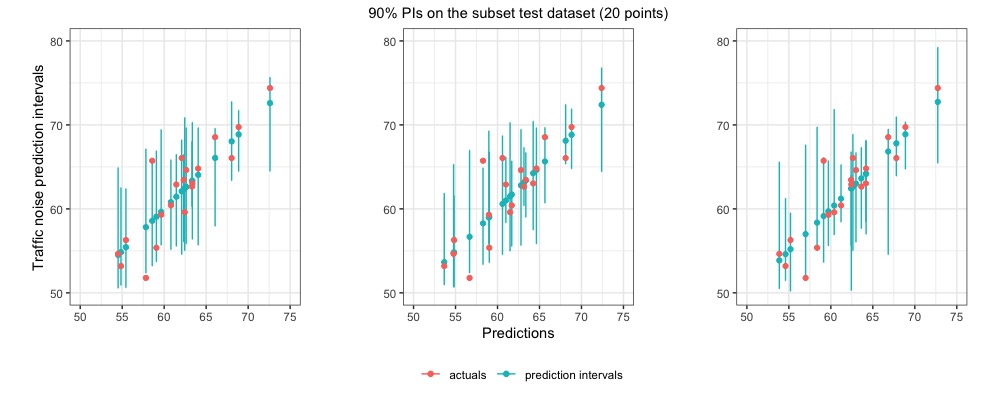}
    \label{fig:pi_noise_20_r}
\caption{Prediction intervals (turquoise) for the traffic noise data from quantile gradient boosting (left), quantile light gradient boosting (center), and quantile extreme gradient boosting (right). The red dots are traffic noise point predictions. The figures on the top row show the prediction intervals for the test set; the figures on the bottom row show the prediction intervals for 20 randomly selected points from the test set.}
\label{fig:noise_pi_1st_r}
\end{figure}

The PIs generated for all three models are shown in Figure \ref{fig:noise_pi_1st_r}. The top row of Figure \ref{fig:noise_pi_1st_r} shows that all three models have wider PIs for lower traffic noise levels, and LightGBM tends to over-predict traffic noise compared to GBM and XGBoost. In terms of the mid-level values of traffic noise, the PIs of QXGBoost cover most of the extremely high and low point predictions whereas quantile GBM and LightGBM do not. This is why the QXGBoost PIs look wider than the Quantile GBM and LightGBM in the middle noise level. One thing to notice for QXGBoost is that it has several very wide PIs at the high end of traffic noise for some observations, whereas the Quantile GBM and LightGBM have relatively consistent lengths of PIs, indicating that quantile GBM and LightGBM might be more stable than the QXGBoost in predicting the high traffic noise levels. Both the first and second rows of the figure \ref{fig:noise_pi_1st_r} show that PIs vary in length, with some being much shorter than others. The relatively wider PIs at both ends of the traffic noise levels are due to a smaller number of observations. Figure \ref{fig:point prediction for noise using 3 models} shows that most of the traffic noise values were distributed between 50 and 75 dB and the ones that are below 50 dB or above 75 dB consisted only 2.9\% and 2.6\% of all the observations, respectively. Thus, it is not unexpected that with less training data, models will generate PIs of high uncertainty with a wider length. 

\begin{figure}
    \centering
    \includegraphics[width=1\textwidth]{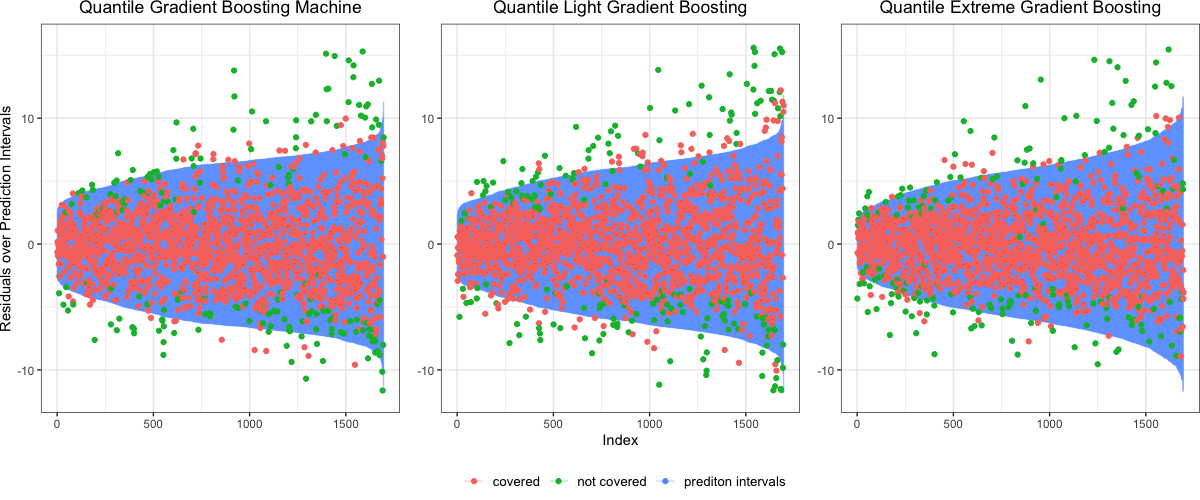}
    \label{fig:pi_ordered_noise_r}
\caption{Ordered and centered PIs for the traffic noise data from all three models. The red dots represent the observations that are covered by PIs. The green dots are the observations that are not covered by PIs. The blue areas show the ordered and centered PIs.}
\label{fig:noise_pi_2ndr}
\end{figure}

\begin{figure}
    \centering
    \includegraphics[width=.6\linewidth,height=.5\textheight,keepaspectratio]{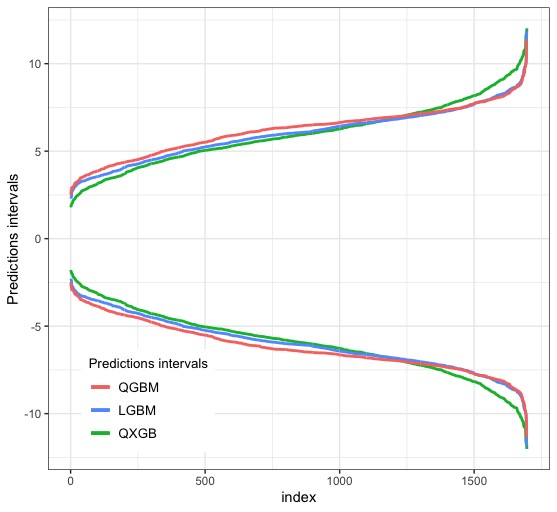}
    \label{fig:pi_overlay_noise_20_r}
\caption{The figure on the second row is the ordered and centered PIs from all three models on the same plot.}
\label{fig:noise_pi_overlay}
\end{figure}

For better visualization, Figure \ref{fig:noise_pi_2ndr} shows all traffic noise observations ordered according to the length of the corresponding PIs. It can be easily seen that the overall width of PIs of the quantile GBM is wider than that of the quantile LightGBM and QXGBoost. QGBoost has a slightly tighter width than the quantile LightGBM. All three models performed relatively well at the middle-level traffic noise and covered almost 90\% of the observed traffic noise in the testing dataset. Figure \ref{fig:noise_pi_overlay} shows the PIs from all three models together for better comparison. Quantile GBM and LightGBM have a wider length of PIs than QXGBoost for most of the observations (around 1300 out of 1649 observations). For the rest of the observations, the PIs width of QXGBoost is wider as was reflected at the high index of figure \ref{fig:noise_pi_overlay}, indicating that the widths of the PIs from QXGBoost vary more than those of quantile GBM and quantile LightGBM.

\section{Conclusion and Discussion}
We developed QXGBoost, a new method that employs quantile regression as the customized objective function for XGBoost for uncertainty analysis. In order to have an objective function that can be differentiable everywhere and have second derivatives, we adapted the concept developed in previous studies for neural networks \cite{chen2007finite, cannon2011quantile} by introducing the Huber norm function to provide a smooth approximation to the true quantile regression. To our knowledge, this is the first study that combines quantile regression, Huber norm function, and XGBoost to do uncertainty analysis for XGBoost. The results from our experiments suggest that QXGBoost produces competitive or better quality PIs compared to quantile GBM and LightGBM. \par 
The primary goal of this study was to develop an easy-to-understand implementation of uncertainty quantification for XGBoost. Even though popular uncertainty quantification methods such as Bayesian or Delta methods for the neural networks have been studied for years \cite{khosravi2011comprehensive}, there are some disadvantages to using them. First, there is a learning curve for non-experts to understand Bayesian or Delta methods. Second, their integration with XGBoost is very challenging due to the fact that tree-based ML methods are fundamentally different from neural network models. GBM was able to use the quantile regression function directly because it does not require second-order approximation for the objective function. LightGBM is similar to XGBoost in terms of its objective optimization method. However, the success of quantile LightGBM requires diving into the source code (c++) and exporting the tree leaf index to the objective function, which would be difficult for people who are not familiar with the source code and programming language. QXGBoost solved the problem simply by defining an optimized objective function. Even though QXGBoost has one additional parameter to tune, it provides a simple and quick solution for performing uncertainty analysis for one of the most powerful ML methods.\par

The additional threshold hyperparameter $\tau$ plays an important role during model fitting. Based on our experiment, $\tau$ controls both the PICP and PINAW of the PIs. Small $\tau$ (less than 1) usually leads to large PICP and PINAW and large $\tau$s typically produce small PICP and PINAW, meaning more constraints on the PIs. In practice, we prefer large PIs coverage with a narrower width. Thus, both the threshold and XGBoost parameters need to be tuned and balanced to achieve better model performance. However, as is shown in figure \ref{fig:huber plot}, as $\tau$ gets larger (more than 2), the Huber smoothed approximation function departs further away from the original quantile regression function, and the poor approximation might lead to biased PIs. In practice, we want to control the threshold parameter $\tau$ under 2. \par

 There are limitations to our method. First, as we have seen from figure \ref{fig:noise_pi_1st_r}, for extremely low and high traffic noise values, when there were few observations for training, QXGBoost could not give informative PIs because less data led to greater uncertainty in the predictions. This is true for other ML models including GBM and LightGBM, which is why all three models have better PI quality for mid-level than at low and high ends of traffic noise. We generally see overestimation for low levels of traffic noise and underestimation for high-level traffic noise. Second, we usually could not achieve the "hard" coverage as the $\tau$ specifies for the proposed QXGBoost method. For example, we have to uniformly pad the quantile intervals by 3\% and 1\% to get a 90\% coverage using QXGBoost for the simulated and traffic noise data, respectively. This limitation is not unique to our method. As is shown in other studies \cite{qrfandbootstrap, qrf}, additional padding was also needed for quantile random forest to achieve desired quantile coverage. Given these limitations, future efforts should be focused on optimizing the objective function of the XGBoost so that the ``hard'' coverage can be achieved without additional padding. \par 

Data uncertainty exists in every stage of research from data collection, modeling, analysis, visualization, and thus drawing conclusions. XGBoost has been widely used in all research areas from image classification to disease diagnosis. The popularity of the XGBoost algorithm motivates the need for its ability to do uncertainty quantification to better understand and capture uncertainty during the modeling stage. Ignoring uncertainty can lead to biased results and thus hinder the improvement of health policies and any other science-based decision making. As the first study proposing an uncertainty method for XGBoost, we fill in the knowledge gap of modelling uncertainty using one of the state-of-art ML algorithms. One important innovation of QXGBoost is its easy implementation and customization. The success of QXGBoost provides potential applications for epidemiological studies, both in modeling exposure uncertainty and assessing health effects. With the increased availability and ubiquity of spatial and health data, uncertainty quantification methods such as QXGBoost can help us better understand the causal effects relationship between exposures and outcomes.

\section*{Acknowledgments}
The research described in this article was conducted under contract to the Health Effects Institute (HEI), an organization jointly funded by the United States Environmental Protection Agency (EPA) (Assistance Award No. CR-83590201) and certain motor vehicle and engine manufacturers. The contents of this article do not necessarily reflect the views of HEI, or its sponsors, nor do they necessarily reflect the views and policies of the EPA or motor vehicle and engine manufacturers. 

\bibliographystyle{unsrt}  
\bibliography{references}

\end{document}